\documentclass[letterpaper, 10 pt, conference]{ieeeconf}
\IEEEoverridecommandlockouts
\usepackage{graphicx}
\usepackage{booktabs}
\usepackage{color, soul}
\usepackage{graphics}
\usepackage{amsmath}
\usepackage{balance}
\usepackage{xcolor}
\usepackage{float}

\title{\LARGE \bf
Low-Burden LLM-Based Preference Learning: Personalizing Assistive Robots from Natural Language Feedback for Users with Paralysis 
}

\author{Keshav Shankar$^{1}$, Dan Ding$^{2}$, and Wei Gao$^{1}$% <-this % stops a space
\thanks{This work was supported by a disability trust requesting anonymity.}% <-this % stops a space
\thanks{$^{1}$Electrical and Computer Engineering, and $^{2}$Rehabilitation Science and Technology, University of Pittsburgh, Pittsburgh, PA. 
        Emails: {\tt\small \{kesha`vshankar, dad5, weigao\}@pitt.edu}}%
}

\begin{document}

\maketitle
\thispagestyle{empty}
\pagestyle{empty}

\begin{abstract}
Physically Assistive Robots require personalized behaviors to ensure user safety and comfort. However, traditional preference learning methods, like exhaustive pairwise comparisons, cause substantial physical and cognitive fatigue for users with severe motor impairments. To solve this, we propose a low-burden, offline framework that translates unstructured natural language feedback directly into deterministic robotic control policies. To safely bridge the gap between ambiguous human speech and robotic code, our pipeline uses Large Language Models (LLMs) grounded in the Occupational Therapy Practice Framework. This clinical reasoning decodes subjective user reactions into explicit physical and psychological needs, which are then mapped into transparent decision trees. Before deployment, an automated ``LLM-as-a-Judge'' verifies the code's structural safety. We validated this system in a simulated meal preparation study with 10 adults with paralysis. Results show our natural language approach significantly reduces user workload compared to traditional baselines. Additionally, occupational therapists confirmed the generated policies are safe and accurately reflect user preferences.
\end{abstract}

\section{Introduction}
Physically Assistive Robots (PARs) assist individuals with motor impairments in daily tasks like object manipulation \cite{nanavati2023physically}. While modern PARs are highly capable of such tasks \cite{kemp2022design}, ensuring user acceptance requires personalization \cite{sorensen2025user}. To maintain safety and comfort during Human-Robot Interaction (HRI), PARs must adapt to each user's specific physical needs and environment. For instance, a user with a spinal cord injury may require the robot to maintain a larger distance to protect their wheelchair and a slower approach speed to feel secure.

These robot behaviors are typically optimized through preference learning \cite{furnkranz2010preference}, utilizing absolute ratings \cite{canal2021preferences} or relative pairwise comparisons \cite{hullermeier2008label} from users. However, traditional preference learning assumes users can provide extensive, iterative feedback. Manually evaluating dozens of robotic configurations causes severe physical fatigue and cognitive overload for individuals with profound motor impairments, such as amyotrophic lateral sclerosis (ALS) \cite{ramirez2008fatigue}. Furthermore, data-driven supervised learning remains infeasible in this domain due to the prohibitive cost of collecting large-scale participant data.

\begin{figure}[t]
    \centering
    \includegraphics[width=\linewidth]{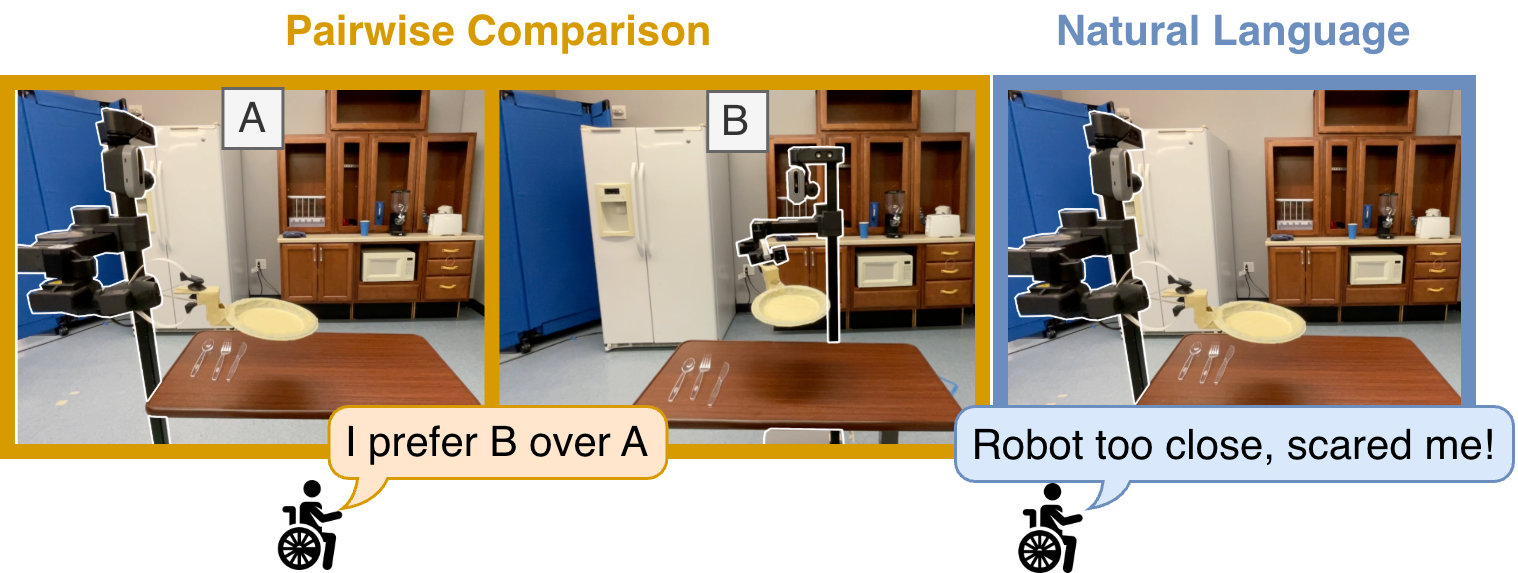}
    \caption{High-burden pairwise comparisons (left) versus our proposed low-burden natural language (right) preferences.}
    \label{fig:methods}
\end{figure}

Large Language Models (LLMs) demonstrate strong capabilities in natural language understanding and reasoning \cite{wei2022chain}, offering an alternative to infer preferences from low-effort natural language feedback, rather than exhaustive pairwise comparisons (Fig. \ref{fig:methods}). Yet, bridging the semantic gap between ambiguous feedback and structured robotic rules remains challenging. Translating a reaction like ``that scared me, it's too fast!'' directly into rigid code (e.g., ``\texttt{IF} robot approaching user, \texttt{THEN} speed = slow'') frequently causes logical errors and misses the user's underlying intent. For individuals with complex physical and psychological requirements, direct translation via standard LLM prompting often proves insufficient. The system must first apply clinical reasoning to contextualize the user's needs, then translate that understanding into explicit, safe behavioral rules across different task states.

In this paper, we propose an offline LLM-based framework that infers interaction preferences for users with paralysis from natural language feedback. Our approach grounds LLM reasoning in the Occupational Therapy Practice Framework (OTPF) \cite{boop2020occupational}, a clinical standard for mapping patient capabilities, systematically decoding ambiguous reactions into core physical and psychological needs. This allows us to reliably translate high-level feedback into interpretable decision tree policies that govern robot behavior. To guarantee safety, we introduce an ``LLM-as-a-Judge'' mechanism that verifies the structural validity of the policies, such as atomic conditions, distinct branching, and closed execution logic, ensuring the PAR safely adapts to the user.

The main contributions of this paper are:
\begin{itemize}
    \item An \textbf{LLM-based framework} integrating clinical domain knowledge (OTPF) to infer user preferences from natural language and generate safe, interpretable decision tree policies.
    \item An \textbf{LLM-as-a-Judge evaluation mechanism} that strictly enforces structural safety and deterministic reliability in the generated policies.
    \item An \textbf{$n=10$ user study} demonstrating our natural language approach significantly reduces physical and cognitive workload compared to traditional preference learning.
\end{itemize}

Through this user study, we explicitly evaluate whether our natural language approach minimizes physical and cognitive burden for paralyzed individuals compared to traditional baselines, while verifying that the framework generates clinically valid, safe robotic behaviors. Specifically, we test two primary hypotheses: \textbf{H1} (\textit{Elicitation Method Burden}), which states that natural language feedback induces significantly less cognitive and physical fatigue for users with motor impairments compared to absolute questionnaires or relative pairwise comparisons; and \textbf{H2} (\textit{System Validity}), which states that the proposed LLM-based pipeline successfully translates unstructured natural language feedback into clinically valid, safe, and personalized robotic behavior policies.

\section{Related Work}

\subsection{Preference Learning}
Preference learning predicts user preferences from their observed feedback \cite{furnkranz2010preference}. Early HRI approaches used absolute ratings (e.g., whether a robot should move slow, medium, or fast) \cite{canal2021preferences}. Because interpretations of terms like ``slow'' vary subjectively, methods shifted to relative pairwise comparisons \cite{bradley1952rank}, where users select their favorite between two options. This approach assumes strict transitivity (i.e., if option A is preferred over option B, and B over C, then A must be preferred over C).

However, human preferences are frequently intransitive and context-dependent \cite{tversky1969intransitivity}. For example, a user might prefer a fast robot over a slow one, and a silent robot over a talkative one, but strongly dislike a fast, silent robot due to startle risks. Label ranking captures these complex cycles without forcing transitivity by treating every comparison as an independent choice \cite{hullermeier2008label}. 

The critical limitation of label ranking is its reliance on exhaustive combinatorial querying. For users with motor impairments, evaluating every possible pair induces severe physical and cognitive fatigue. Even state-of-the-art active preference learning, which dynamically queries only the most mathematically informative pairs \cite{sadigh2017active}, still demands an unacceptable volume of interaction. This highlights a critical need for low-effort elicitation methods.

\subsection{Clinical Reasoning in Assistive Care}
Understanding the nuanced needs of paralyzed users requires specialized domain knowledge, standardized by the OTPF \cite{boop2020occupational}. The OTPF categorizes the impact of disabilities on daily living into the \emph{Domain} (i.e., client factors, occupations, contexts, performance skills, performance patterns) and the \emph{Process} (i.e., evaluation, intervention, outcomes). Clinicians evaluate users by creating an Occupational Profile and analyzing their Occupational Performance to identify specific interaction barriers.

Crucially, while traditional medical models use hypothetico-deductive reasoning to diagnose physical deficits \cite{elstein1978medical}, occupational therapy emphasizes narrative reasoning \cite{mattingly1998search}. Narrative reasoning is cumulative and subjective, seeking the ``Why'' behind an activity from the user's perspective. For instance, narrative reasoning reveals whether a user desires a slower robot due to a physical motor limitation or a psychological need to feel safe. 

\subsection{Large Language Models}
LLMs demonstrate strong zero-shot reasoning, particularly via Chain-of-Thought (CoT) prompting \cite{wei2022chain}. In preference learning, LLMs currently improve efficiency by selecting informative questions \cite{li2023eliciting}. In robotics, LLMs act as high-level semantic planners \cite{huang2022language} for existing Application Programming Interfaces (APIs), or as Vision-Language-Action (VLA) models outputting direct motor commands \cite{zitkovich2023rt}. However, VLAs lack the transparency required for safety-critical clinical environments.

Furthermore, standard LLMs frequently suffer from ``semantic drift'' over long context windows, conflating variables or losing track of prior states \cite{liu2024lost}. To maintain logical consistency, prompt chaining decomposes complex reasoning into explicit steps \cite{wu2022ai}. Because safety-critical robotic logic must be perfectly deterministic, natural language must be semantically parsed \cite{warren1982efficient} into structured formats like decision trees \cite{freitas2014comprehensible}.

Since LLM generation is non-deterministic, standard reliability approaches query the model multiple times and apply majority voting \cite{wang2022self}. However, exact-match voting struggles with complex topological structures like decision trees \cite{weinberg2019selecting}. Furthermore, the semantic conditions inside natural language tree nodes are difficult to mathematically verify. Consequently, using a highly capable model to score outputs, an approach known as ``LLM-as-a-Judge'' \cite{zheng2023judging}, has emerged as a robust standard, including in clinical domains \cite{croxford2025current}. Synthesizing these components to translate unstructured clinical feedback into verifiable robotic policies remains an unexplored gap in HRI, which our framework directly addresses.

\section{Methodology}

\begin{figure*}[t]
    \vspace{2mm}
    \centering
    \includegraphics[width=\textwidth]{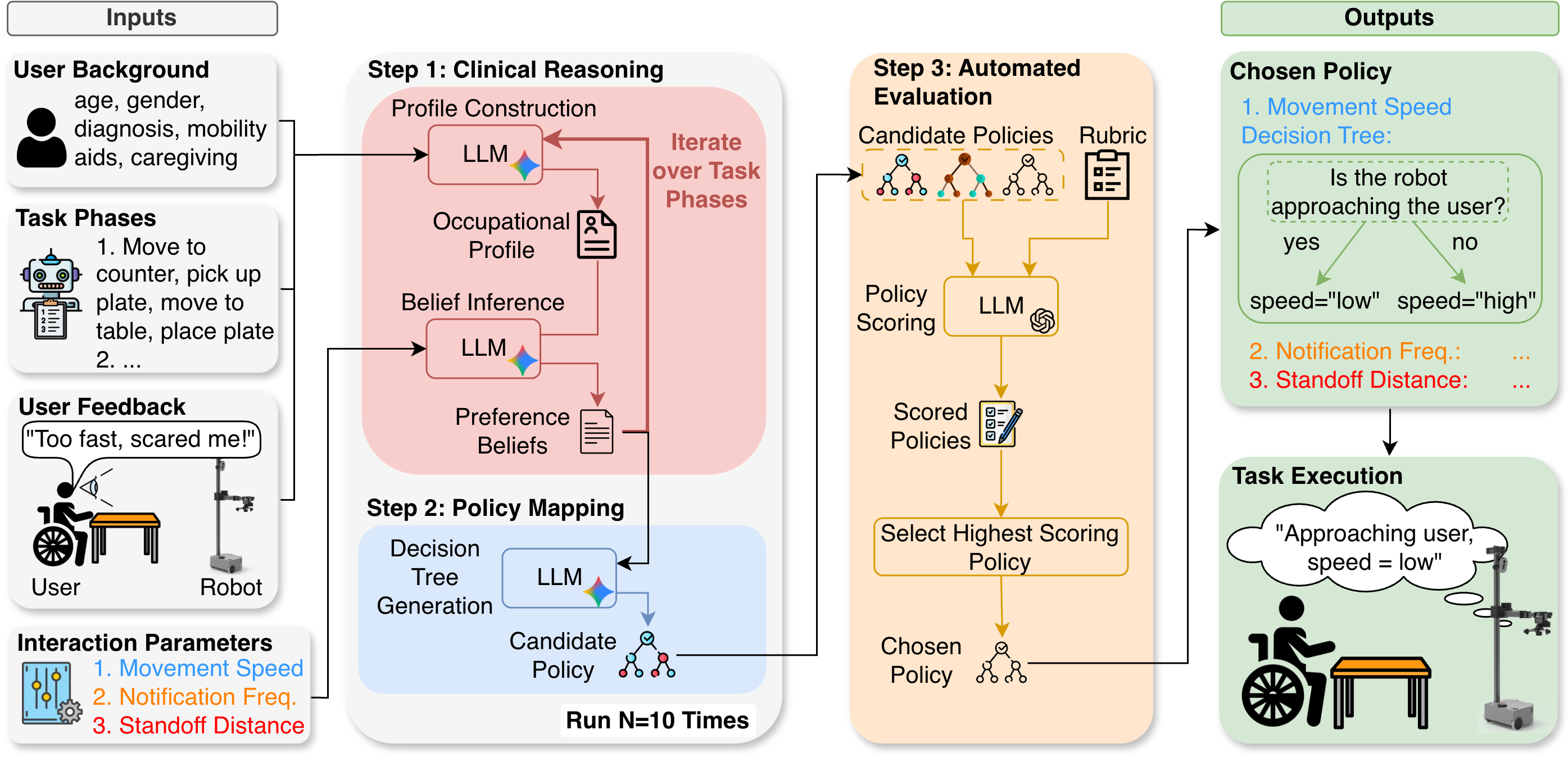}
    \caption{Proposed pipeline for translating unstructured natural language feedback into verifiable robotic policies.}
    \label{fig:framework}
\end{figure*}

The proposed framework (Fig. \ref{fig:framework}) infers a user's preferences from their unstructured natural language feedback, translating them into deterministic robotic policies through three sequential steps: \textit{Clinical Reasoning} to extract preferences, \textit{Policy Mapping} to format logic, and \textit{Automated Evaluation} to ensure structural validity. 

\subsection{Pipeline Inputs}
To ground the LLM's reasoning, the system initializes with contextual data, $C = \langle \mathcal{B}, \mathcal{T}, \mathcal{P}, \mathcal{F} \rangle$:

\begin{itemize}
    \item \textbf{User Background ($\mathcal{B}$):} Clinical details regarding the user's capabilities and context (e.g., diagnosis, mobility aids, technology input methods), gathered via a standardized self-report intake interview.
    \item \textbf{Task Phases ($\mathcal{T}$):} Sequential text descriptions of the robotic task, $\mathcal{T} = \{t_1, t_2, \dots, t_k\}$ (e.g., 1. Drive to counter, 2. Pick up plate).
    \item \textbf{Interaction Parameters ($\mathcal{P}$):} Configurable robot behaviors and their allowable discrete states, $\mathcal{P} = \{p_1, \dots, p_m\}$ (e.g., Speed: low/medium/high).
    \item \textbf{User Feedback ($\mathcal{F}$):} Transcribed natural language feedback provided by the user after observing the robot execute a task under default settings.
\end{itemize}

\subsection{Step 1: Clinical Reasoning}
To systematically interpret feedback ($\mathcal{F}$) within the task context, the pipeline applies the OTPF \cite{boop2020occupational} to deconstruct user reactions into physical and psychological needs. Rather than directly mapping language to code, the LLM first categorizes $\mathcal{B}$ and $\mathcal{F}$ into formal clinical domains: Client Factors (e.g., specific muscle weakness), Contexts (e.g., physical environment), and Performance Patterns (e.g., habits or routines). This grounds narrative reasoning in verifiable clinical logic before adjusting any robot parameters.

\begin{figure*}[t]
    \vspace{2mm}
    \centering
    \includegraphics[width=\textwidth]{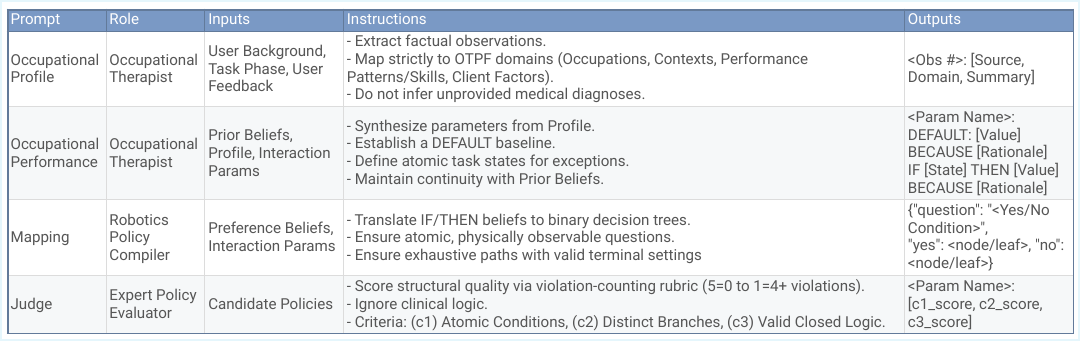}
    \caption{Overview of the LLM prompt structures, including roles, inputs, instructions, and outputs for the pipeline.}
    \label{fig:prompts}
\end{figure*}

Because feedback often references specific moments, this step utilizes prompt chaining via Google's Gemini 2.5 Flash to iterate sequentially over each task phase $t_i \in \mathcal{T}$:
\begin{enumerate}
    \item \textbf{Profile Construction:} The LLM synthesizes $\mathcal{B}$, $t_i$, and $\mathcal{F}$ to generate a localized \textit{Occupational Profile} (Fig. \ref{fig:prompts}), classifying feedback into exact, relevant OTPF domains (e.g., mapping ``I don't like it on my left side, it makes me feel unsafe'' directly to \textit{Client Factors: Values and Beliefs}).
    \item \textbf{Belief Inference:} Using the profile and prior phase insights, the LLM infers explicit preference beliefs (\textit{Occupational Performance}) (Fig. \ref{fig:prompts}). These define a \texttt{DEFAULT} parameter state, \texttt{IF/THEN} exceptions, and a clinical \texttt{BECAUSE} justification (e.g., ``\texttt{DEFAULT} Standoff Distance = medium; \texttt{IF} approaching user side, \texttt{THEN} Standoff Distance = high \texttt{BECAUSE} user expressed discomfort with robot entering personal space due to stroke (Client Factors, Performance Patterns)'').
\end{enumerate}

This process iterates through all task phases to produce a single, comprehensive list of clinically grounded beliefs.

\subsection{Step 2: Policy Mapping}
To convert semantic beliefs into executable control, the LLM translates the final belief list into a \textit{Policy Set}, $\Pi$, via \textit{Mapping} (Fig. \ref{fig:prompts}). A policy set consists of explicit decision trees formatted as JSON objects, with one distinct tree $\pi_j$ for each interaction parameter $p_j \in \mathcal{P}$. The LLM maps the \texttt{IF/THEN} clauses into binary Yes/No question nodes.

Translating natural language into rigid trees intentionally decouples high-level semantic reasoning from low-level physical execution. A standard robotic task planner can parse these discrete JSON structures to invoke existing APIs, guaranteeing the robot's movements remain strictly bounded by safe kinematic limits. This step also utilizes Gemini 2.5 Flash.

\subsection{Step 3: Automated Evaluation}
Because LLMs exhibit non-deterministic generation and can introduce syntactic or logical hallucinations, the system repeats the pipeline $N=10$ times to generate a diverse candidate pool of policy sets $\mathbf{C} = \{\Pi^{(1)}, \dots, \Pi^{(N)}\}$. This sample size provides a sufficient statistical spread of alternative syntactic structures. To select the safest candidate, an LLM \textit{Judge} mechanism evaluates them (Fig. \ref{fig:prompts}). We utilize OpenAI's GPT-5.1 to prevent preference leakage (i.e., evaluation bias resulting from a model, or a model from the same family, scoring its own outputs) \cite{li2025preference}.

Crucially, the Judge evaluates structural validity rather than subjective quality. Each decision tree $\pi_j^{(n)}$ is scored against three standard criteria \cite{breiman2017classification}: Atomic Conditions ($c_1$, each node evaluates exactly one observable state variable); Distinct Branches ($c_2$, explicit True/False paths); and Valid Closed Logic ($c_3$, all topological paths terminate in a predefined valid state).

Using a violation-counting rubric, the Judge assigns a score from 5 (0 violations) to 1 ($\ge 4$ violations) per criterion. A total structural score $S$ is calculated for the policy set $\Pi^{(n)}$ by summing the criteria scores $c \in \mathcal{C}$ (where $\mathcal{C} = \{c_1, c_2, c_3\}$) across all parameter trees $\pi_j$:

\begin{equation} \label{eq:score}
S(\Pi^{(n)}) = \sum_{j=1}^{m} \sum_{c \in \mathcal{C}} \text{Judge}(\pi_j^{(n)}, c)
\end{equation}

The policy set $\Pi^*$ with the highest aggregate score is selected. Ties are broken by selecting the earliest generated set:
\begin{equation} \label{eq:argmax}
\Pi^* = \underset{\Pi^{(n)} \in \mathbf{C}}{\arg\max} \, S(\Pi^{(n)}).
\end{equation}

The specific decision trees within this winning policy set $\Pi^*$ can then be reliably deployed to the assistive robot. If all $N=10$ generated candidates fail to achieve a perfect structural score, the system halts and flags the policy for human review.

\section{Experimental Validation}

\subsection{Study Goals and Hypotheses}
As outlined in the Introduction, this study evaluates methods for eliciting robot interaction preferences, specifically movement speed, notification frequency, and standoff distance \cite{canal2017taxonomy, akalin2023taxonomy}, under the core hypotheses of reducing elicitation burden (\textbf{H1}) and maintaining system validity (\textbf{H2}).

\subsection{Task and Setup}

\subsubsection{Task Design \& Environment}
\begin{figure}[b]
    \centering
    \includegraphics[width=\columnwidth]{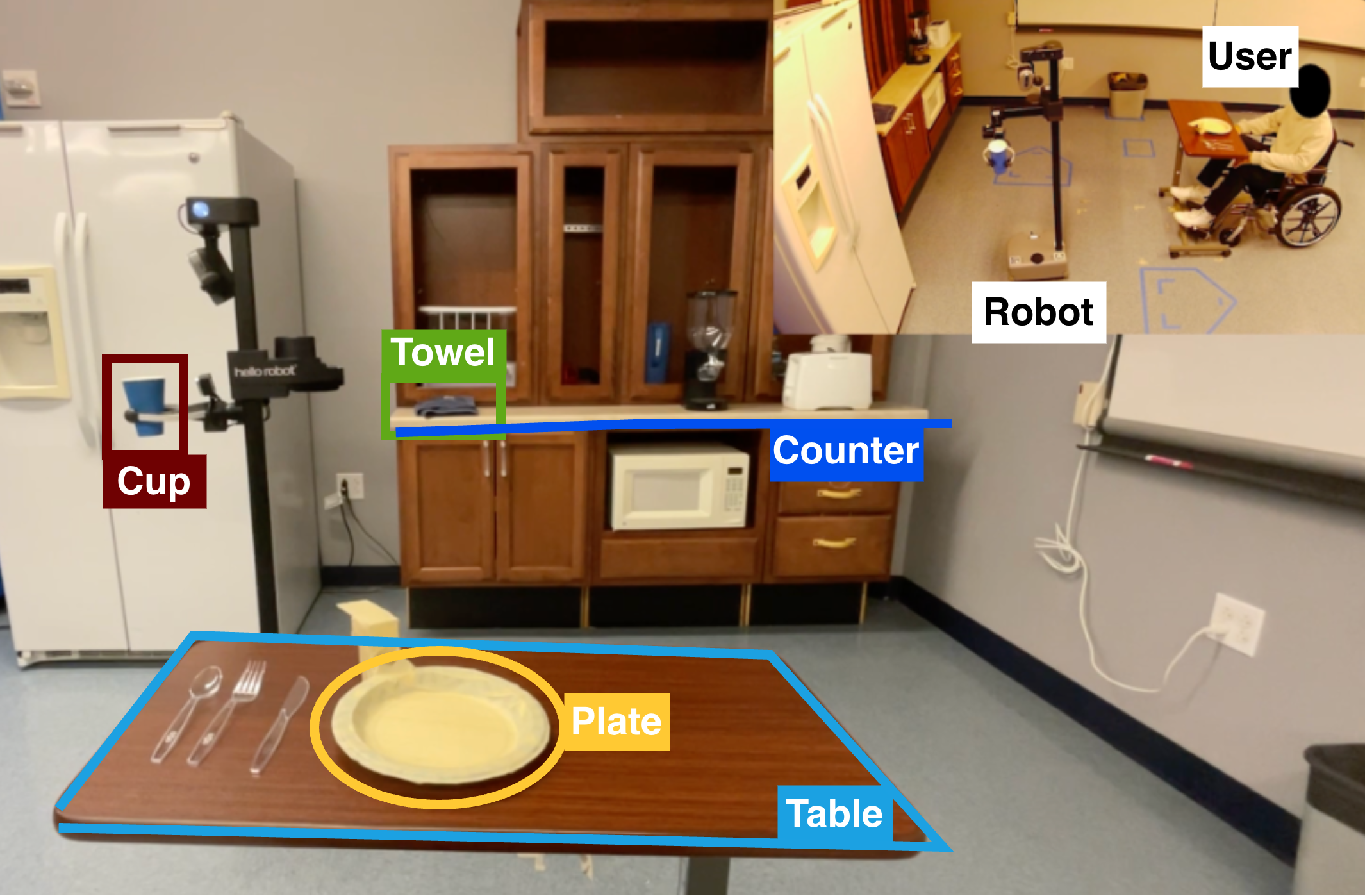}
    \caption{Simulated kitchen environment with target objects, user, and robot.}
    \label{fig:environment}
\end{figure}

We designed a simulated pick-and-place meal preparation task \cite{nanavati2023physically} in a wheelchair-accessible kitchen using a Hello Robot Stretch 3 mobile manipulator \cite{kemp2022design} (Fig. \ref{fig:environment}). The task required the robot to retrieve objects and approach the user to place them. To focus strictly on high-level proximity and signaling preferences, we intentionally excluded complex physical handovers and feeding \cite{koay2014social, choi2009hand}. Furthermore, to evaluate these preferences while assuming ideal technological capability \cite{steinfeld2009oz}, robot manipulation was controlled via Wizard-of-Oz teleoperation for consistent execution, while navigation and communication remained autonomous. Because our target demographic frequently relies on caregivers for similar object-retrieval tasks, they possess established mental models enabling them to provide grounded feedback \cite{norman2014some}.

\subsubsection{Participants}
With IRB approval from the University of Pittsburgh (STUDY26010079), we recruited 10 adults ($M=45.0$ years, $SD=14.7$; 7 females, 3 males) through an existing institutional physical disability registry. Inclusion criteria required participants to be $\ge 18$ years of age, hold a high school diploma or equivalent credential, and have a clinical diagnosis resulting in partial or complete paralysis (e.g., ALS, spinal cord injury, stroke) with upper extremity limitations impacting daily activities. Eligible individuals were also required to be proficient in English, capable of viewing video stimuli on a computer or tablet (independently or via assistive technology), and comfortable providing feedback using speech, text, or an alternative accessible input method. As detailed in Table \ref{tab:fim_summary}, participants exhibited severe physical impairments (Mean Motor Functional Independence Measure (FIM) \cite{kidd1995functional}: 48.1/91), but possessed the requisite cognitive and communicative capacity to evaluate the interaction tasks (Mean Cognitive FIM: 33.9/35). Baseline comfort with technology was high ($M=4.4/5, SD=0.7$), quantified via a 5-point self-report scale during intake.

\begin{table}[t]
    \vspace{2mm}
    \centering
    \caption{Aggregate FIM Scores}
    \label{tab:fim_summary}
    \begin{tabular}{lccc}
    \toprule
    \textbf{FIM Category} & \textbf{Average} & \textbf{Max} & \textbf{Std. Dev.} \\
    \midrule
    \multicolumn{4}{l}{\textit{Motor Domain}} \\
    \hspace{3mm} Self-care & 22.4 & 42 & 11.7 \\
    \hspace{3mm} Sphincter Control & 8.5 & 14 & 4.7 \\
    \hspace{3mm} Transfers & 11.0 & 21 & 7.4 \\
    \hspace{3mm} Locomotion & 6.2 & 14 & 3.1 \\
    \midrule
    \multicolumn{4}{l}{\textit{Cognitive Domain}} \\
    \hspace{3mm} Communication & 13.5 & 14 & 1.1 \\
    \hspace{3mm} Social Cognition & 20.4 & 21 & 1.0 \\
    \bottomrule
    \end{tabular}
\end{table}

\subsection{Evaluation A: Elicitation Method Burden (Testing H1)}

\subsubsection{Protocol \& Metrics}
Using a within-subjects design, participants observed pre-recorded videos of the robot from first-person and overhead perspectives (Fig. \ref{fig:environment}). Recent literature confirms that as long as the robot's task is relatively simple, watching pre-recorded videos of a robot produces similar subjective user feedback to interacting with a physical robot in person \cite{woods2006methodological, esterwood2025virtually}. To mitigate demand characteristics \cite{irfan2018social}, the order of methods shown was randomized using a Latin Square design, and participants were explicitly informed the robot ran on pre-recorded logic. To quantify user burden, we measured Total Time on Task and subjective cognitive workload via the NASA-TLX \cite{hart1988development} across three elicitation methods:
\begin{itemize}
    \item \textbf{Method 1 (Baseline A - Questionnaire):} Users explicitly selected parameter levels across 7 segmented contexts via a standard absolute-rating questionnaire.
    \item \textbf{Method 2 (Baseline B - Pairwise):} Users evaluated exhaustive pairs of configurations (Label Ranking \cite{hullermeier2008label}). Complete coverage required up to $\frac{m(m-1)}{2}$ queries per context, necessitating 153 comparisons. A 30-minute hard stop was enforced to prevent extreme fatigue.
    \item \textbf{Method 3 (Proposed - Natural Language):} Users viewed the full task in phases and provided open-ended natural language feedback (speech or text).
\end{itemize}

\begin{figure}[t]
    \vspace{2mm}
    \centering
    \includegraphics[width=\linewidth]{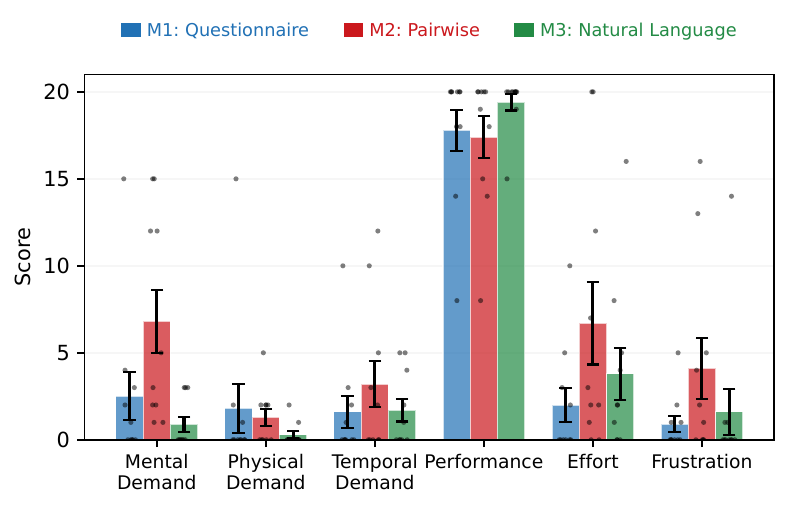}
    \caption{Distribution of NASA-TLX subscale scores across the three elicitation methods: questionnaire (M1), pairwise comparison (M2), and natural language feedback (M3). Solid bars represent mean scores, error bars denote the standard error of the mean, and black dots indicate individual participant scores. Lower scores indicate lower perceived user burden. Scoring for the Performance subscale is inverted.}
    \label{fig:nasa_tlx}
\end{figure}

\subsubsection{Results}
Participants completed Method 1 in 4 minutes 10 seconds on average, and Method 3 in 6 minutes 16 seconds. Method 3 easily accommodated highly variable feedback lengths (4 to 226 words; $M=73.0, SD=55.2$). Conversely, Method 2 induced severe fatigue; participants completed only $41 \pm 2$ comparisons before reaching the 30-minute limit, which was near the minimum threshold required to establish an intransitive model.

Method 3 substantially reduced perceived burden across all NASA-TLX subscales compared to Method 2 (Fig. \ref{fig:nasa_tlx}). Notably, despite requiring roughly two more minutes to complete than Method 1, Method 3 yielded lower Mental and Physical Demand and higher perceived Performance. While the aggregate means for Effort and Frustration in Method 3 appear marginally higher than Method 1, this variance was heavily skewed by a single outlier who experienced severe fatigue spillover after being randomized to complete Method 2 immediately prior. Excluding this participant, Method 3's Effort and Frustration actually fell below Method 1.

\subsection{Evaluation B: System Validation (Testing H2)}
Feedback from Method 3 was processed through our proposed LLM pipeline to evaluate internal module reliability and end-to-end policy alignment.

\subsubsection{Internal Module Validation}
\begin{figure}[t]
    \vspace{2mm}
    \centering
    \includegraphics[width=\linewidth]{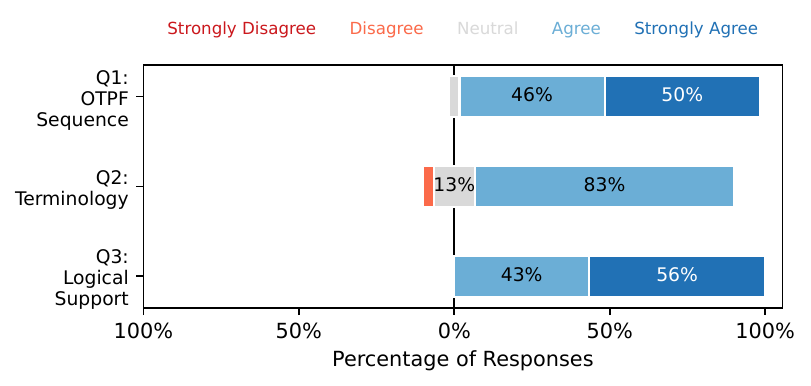}
    \caption{Independent clinician validation of the system's reasoning traces on a 5-point Likert scale. Evaluators assessed adherence to the OTPF evaluation sequence (Q1), correct application of domain terminology (Q2), and logical grounding in raw user feedback (Q3).}
    \label{fig:reasoning_clinician}
\end{figure}

\begin{figure}[t]
    \centering
    \includegraphics[width=0.5\textwidth]{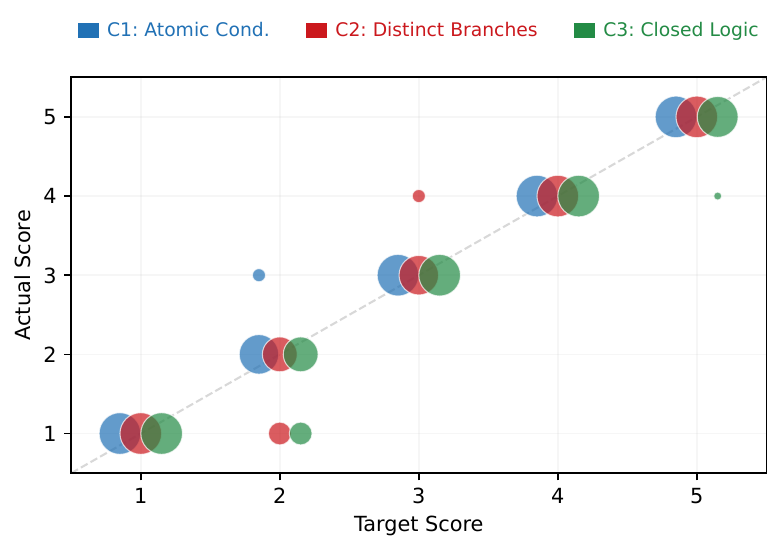}
    \caption{Performance of the LLM Judge on 360 structurally perturbed policies. Actual assigned scores are plotted against target scores (injected violations) across three structural evaluation criteria: Atomic Conditions (C1), Distinct Branches (C2), and Valid Closed Logic (C3).}
    \label{fig:llm_judge_ablation}
\end{figure}

We first verified that the intermediate artifacts generated by the pipeline were structurally and clinically sound:
\begin{itemize}
    \item \textbf{Reasoning Module:} Three blinded occupational therapists ($M=6$ years experience) evaluated the LLM's clinical reasoning traces across 3 distinct evaluation items for all 10 participants (Fig. \ref{fig:reasoning_clinician}). To evaluate inter-rater reliability, we computed the overall adjacent agreement (i.e., percentage of items where all three clinicians' ratings fell within 1 point of each other on the 5-point scale), yielding a high reliability of 93.3\% across all 30 evaluation instances (10 participants $\times$ 3 items). Clinicians noted minor deductions stemming from terminology misclassifications (e.g., labeling robotic hardware traits as an ``environmental factor'') and a lack of proactive clinical foresight (e.g., preemptively adjusting parameters purely on diagnosis before explicit user feedback).
    \item \textbf{Mapping Module:} Because text-overlap metrics \cite{laban2022summac} fail on discrete JSON logic \cite{van2024field}, two domain experts manually traced all 42 clinical beliefs, across the 10 participants, to the generated decision trees to verify there were no hallucinations. The generated decision trees achieved a 100\% (42/42) constraint satisfaction rate, proving safe translation from clinical reasoning to deterministic logic.
    \item \textbf{LLM Judge:} We systematically injected varying levels of structural violations into the final, clinically approved decision trees. This generated 360 perturbed policies representing all combinations of targeted errors across criteria $c_1, c_2,$ and $c_3$. The Judge accurately identified and proportionally penalized these injected faults (Fig. \ref{fig:llm_judge_ablation}) with 96.94\% accuracy (MAE = 0.0306), confirming its reliability.
\end{itemize}

\subsubsection{End-to-End Policy Alignment}
\begin{figure}[t]
    \vspace{2mm}
    \centering
    \includegraphics[width=\linewidth]{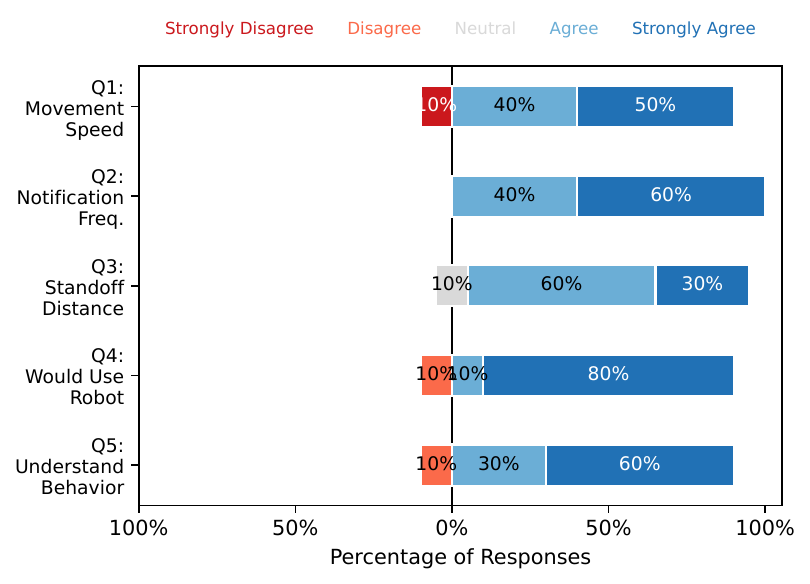}
    \caption{User evaluation of end-to-end policy alignment on a 5-point Likert scale. Users assessed the appropriateness of derived movement speed (Q1), notification frequency (Q2), and standoff distance (Q3), as well as their willingness to use the robot (Q4) and clear understanding of its behavior (Q5).}
    \label{fig:mapping_user}
\end{figure}

\begin{figure}[t]
    \centering
    \includegraphics[width=\linewidth]{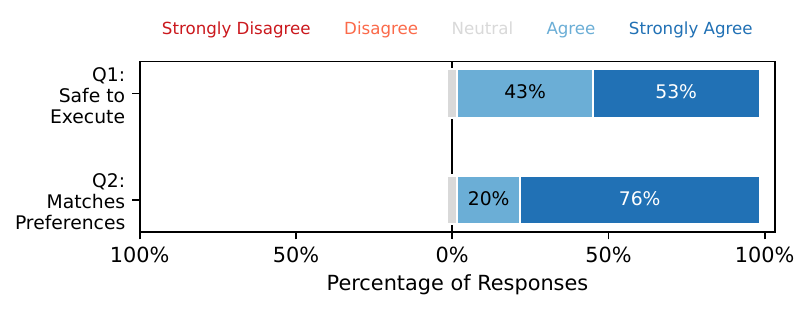}
    \caption{Independent clinical safety check of the mapped robot policies on a 5-point Likert scale. Clinicians evaluated physical/practical risk avoidance (Q1) and accurate alignment with the user's explicit preferences (Q2).}
    \label{fig:mapping_clinician}
\end{figure}

We evaluated the final generated decision tree policies for subjective user satisfaction and clinical safety.
\begin{itemize}
    \item \textbf{User Validation:} Participants reviewed the robotic policies generated from their own feedback across 5 evaluation questions (Fig. \ref{fig:mapping_user}). The evaluation demonstrated strong consistency in user sentiment, achieving an 86.7\% overall inter-user adjacent agreement (calculated as the mean pairwise agreement within 1 point across all questions). Users overwhelmingly agreed that the derived parameters were appropriate, reporting a clear understanding of the robot's planned behavior and a high willingness to use the robot under these personalized policies.
    \item \textbf{Clinical Safety Check:} The same three blinded clinicians reviewed the final mapped policies across all 10 participant cases (Fig. \ref{fig:mapping_clinician}). High clinical alignment was achieved, with a 90.0\% adjacent agreement across the evaluations, strongly confirming that the generated behaviors accurately reflected explicit user preferences without posing physical or practical risks. Clinicians noted only that the system lacked dynamic adaptation (e.g., automatically lowering notification frequency over time as the user's comfort builds, without explicit request).
\end{itemize}

\section{Discussion}

\subsection{Interpretation}
The results demonstrate that natural language (NL) feedback is a viable, low-burden alternative to exhaustive pairwise comparisons. While absolute questionnaires (Method 1) also induced low burden in our specific task, they inherently lack scalability, requiring researchers to manually curate rigid surveys for every new robot capability or context. Conversely, NL scales effortlessly, allowing users to address parameters indirectly and organically.

Regarding validity, grounding LLM reasoning in the OTPF and applying an automated structural judge reliably extracts deterministic policies while strictly mitigating logical hallucinations. We do not claim these policies are mathematically more optimal than exhaustive methods; rather, we establish the feasibility of safe, low-burden personalization. Furthermore, by explicitly decoupling preference extraction from low-level control, the framework is theoretically hardware-agnostic. It can scale to other interaction parameters, different robotic platforms, and similar low-complexity pick-and-place tasks, provided the target system possesses the functional capability to execute the parameterized behaviors.

\subsection{Limitations}
This exploratory study has several key limitations. First, while a sample size of $n=10$ is typical and standard for HRI co-design evaluations with specialized, hard-to-recruit clinical populations (such as individuals with paralysis), it inherently restricts the ability to draw broad statistical generalizations across the highly variable target demographic.

Second, regarding input modalities, while the LLM demonstrated robustness to minor Automatic Speech Recognition (ASR) errors from participants with mild dysarthria, we only tested users capable of speaking or typing. For severe cases, communication relies on noisy or limited Augmentative and Alternative Communication modalities, where input constraints or ASR failures could propagate errors through the pipeline.

Third, our framework relies on sequential LLM modules (clinical reasoning, policy mapping). While our internal evaluation confirms the accuracy of individual stages, we did not perform a formal system ablation. However, our architecture design was explicitly motivated by the fact that directly mapping raw, ambiguous natural language feedback to deterministic decision trees without an intermediate clinical reasoning layer introduced severe logical errors during initial pilot testing.

Fourth, the system operates offline to establish a highly stable behavioral baseline sufficient for routine tasks, meaning it lacks continuous online adaptation or real-time learning over long-term deployment. While underlying hardware limits fundamentally guarantee low-level physical safety and collision constraints, evaluating the generated decision trees for psychological comfort in dynamic, contact-rich Activities of Daily Living (ADLs), such as assisted feeding, remains an open area for multi-scenario validation.

Fifth, we did not quantitatively benchmark the final executed behaviors against baseline policies from Methods 1 and 2. A direct mathematical comparison was precluded because severe cognitive and physical fatigue caused premature termination during the exhaustive pairwise trials, leaving the baseline models incomplete. Furthermore, our natural language framework generates higher-level, generalized behavioral rules, whereas questionnaires yield highly rigid, context-specific values.

\subsection{Future Work}
Natural extensions of this research will transition the framework into a multimodal, adaptive system. To account for ASR limitations with severe dysarthria, future iterations should incorporate alternative input modalities, such as vision-based interpretation of contextual cues (e.g., facial expressions of discomfort). To overcome the lack of inferred habituation without forcing users to restart the entire learning process, subsequent work will explore adaptive update modules capable of continuously calibrating existing rules via brief, ongoing feedback. Finally, deploying these policies onto a physical robot during live ADLs will enable formal benchmarking and evaluate system adaptation to open-ended, real-world requests.

\section{Conclusion}
In this paper, we presented a novel, offline framework to personalize Physically Assistive Robots for individuals with severe motor impairments using unstructured natural language. By replacing exhaustive pairwise comparisons with a clinically grounded Large Language Model pipeline, we demonstrated that safe, deterministic personalization can be achieved while drastically reducing physical fatigue and cognitive burden. Crucially, our automated structural validation mechanism ensures that the inherent non-determinism of large language models does not compromise the strict safety requirements of human-robot interaction. While current limitations include a reliance on vocal speech and the need for explicit user prompts to generate long-range dynamic policies, this work establishes a highly scalable foundation for accessible preference elicitation. Future work will deploy this framework onto physical robots during dynamic daily activities, incorporating multimodal inputs to continuously and safely adapt to evolving user needs.

%\addtolength{\textheight}{-12cm}

\balance

\bibliographystyle{IEEEtran}
\bibliography{references}

\begin{thebibliography}{10}
\providecommand{\url}[1]{#1}
\csname url@samestyle\endcsname
\providecommand{\newblock}{\relax}
\providecommand{\bibinfo}[2]{#2}
\providecommand{\BIBentrySTDinterwordspacing}{\spaceskip=0pt\relax}
\providecommand{\BIBentryALTinterwordstretchfactor}{4}
\providecommand{\BIBentryALTinterwordspacing}{\spaceskip=\fontdimen2\font plus
\BIBentryALTinterwordstretchfactor\fontdimen3\font minus \fontdimen4\font\relax}
\providecommand{\BIBforeignlanguage}[2]{{%
\expandafter\ifx\csname l@#1\endcsname\relax
\typeout{** WARNING: IEEEtran.bst: No hyphenation pattern has been}%
\typeout{** loaded for the language `#1'. Using the pattern for}%
\typeout{** the default language instead.}%
\else
\language=\csname l@#1\endcsname
\fi
#2}}
\providecommand{\BIBdecl}{\relax}
\BIBdecl

\bibitem{nanavati2023physically}
A.~Nanavati, V.~Ranganeni, and M.~Cakmak, ``Physically assistive robots: A systematic review of mobile and manipulator robots that physically assist people with disabilities,'' \emph{Annual Review of Control, Robotics, and Autonomous Systems}, vol.~7, 2023.

\bibitem{kemp2022design}
C.~C. Kemp, A.~Edsinger, H.~M. Clever, and B.~Matulevich, ``The design of stretch: A compact, lightweight mobile manipulator for indoor human environments,'' in \emph{2022 International Conference on Robotics and Automation (ICRA)}.\hskip 1em plus 0.5em minus 0.4em\relax IEEE, 2022, pp. 3150--3157.

\bibitem{sorensen2025user}
L.~S{\o}rensen, D.~T.~S. Johannesen, H.~Melkas, and H.~M. Johnsen, ``User acceptance of a home robotic assistant for individuals with physical disabilities: explorative qualitative study,'' \emph{JMIR Rehabilitation and Assistive Technologies}, vol.~12, no.~1, p. e63641, 2025.

\bibitem{furnkranz2010preference}
J.~F{\"u}rnkranz and E.~H{\"u}llermeier, ``Preference learning and ranking by pairwise comparison,'' in \emph{Preference learning}.\hskip 1em plus 0.5em minus 0.4em\relax Springer, 2010, pp. 65--82.

\bibitem{canal2021preferences}
G.~Canal, C.~Torras, and G.~Aleny{\`a}, ``Are preferences useful for better assistance? a physically assistive robotics user study,'' \emph{ACM Transactions on Human-Robot Interaction (THRI)}, vol.~10, no.~4, pp. 1--19, 2021.

\bibitem{hullermeier2008label}
E.~H{\"u}llermeier, J.~F{\"u}rnkranz, W.~Cheng, and K.~Brinker, ``Label ranking by learning pairwise preferences,'' \emph{Artificial Intelligence}, vol. 172, no. 16-17, pp. 1897--1916, 2008.

\bibitem{ramirez2008fatigue}
C.~Ramirez, M.~E. Pimentel~Piemonte, D.~Callegaro, and H.~C. Almeida Da~Silva, ``Fatigue in amyotrophic lateral sclerosis: frequency and associated factors,'' \emph{Amyotrophic Lateral Sclerosis}, vol.~9, no.~2, pp. 75--80, 2008.

\bibitem{wei2022chain}
J.~Wei, X.~Wang, D.~Schuurmans, M.~Bosma, F.~Xia, E.~Chi, Q.~V. Le, D.~Zhou \emph{et~al.}, ``Chain-of-thought prompting elicits reasoning in large language models,'' \emph{Advances in neural information processing systems}, vol.~35, pp. 24\,824--24\,837, 2022.

\bibitem{boop2020occupational}
C.~Boop, S.~M. Cahill, C.~Davis, J.~Dorsey, V.~Gibbs, B.~Herr, K.~Kearney, L.~Metzger, J.~Miller, A.~Owens \emph{et~al.}, ``Occupational therapy practice framework: Domain and process fourth edition,'' \emph{AJOT: American Journal of Occupational Therapy}, vol.~74, no.~S2, pp. 1--85, 2020.

\bibitem{bradley1952rank}
R.~A. Bradley and M.~E. Terry, ``Rank analysis of incomplete block designs: I. the method of paired comparisons,'' \emph{Biometrika}, vol.~39, no. 3/4, pp. 324--345, 1952.

\bibitem{tversky1969intransitivity}
A.~Tversky, ``Intransitivity of preferences.'' \emph{Psychological review}, vol.~76, no.~1, p.~31, 1969.

\bibitem{sadigh2017active}
D.~Sadigh, A.~Dragan, S.~Sastry, and S.~Seshia, ``Active preference-based learning of reward functions,'' 2017.

\bibitem{elstein1978medical}
A.~S. Elstein, L.~S. Shulman, and S.~A. Sprafka, \emph{Medical problem solving: An analysis of clinical reasoning}.\hskip 1em plus 0.5em minus 0.4em\relax Harvard University Press, 1978.

\bibitem{mattingly1998search}
C.~Mattingly, ``In search of the good: Narrative reasoning in clinical practice,'' \emph{Medical anthropology quarterly}, vol.~12, no.~3, pp. 273--297, 1998.

\bibitem{li2023eliciting}
B.~Z. Li, A.~Tamkin, N.~Goodman, and J.~Andreas, ``Eliciting human preferences with language models,'' \emph{arXiv preprint arXiv:2310.11589}, 2023.

\bibitem{huang2022language}
W.~Huang, P.~Abbeel, D.~Pathak, and I.~Mordatch, ``Language models as zero-shot planners: Extracting actionable knowledge for embodied agents,'' in \emph{International conference on machine learning}.\hskip 1em plus 0.5em minus 0.4em\relax PMLR, 2022, pp. 9118--9147.

\bibitem{zitkovich2023rt}
B.~Zitkovich, T.~Yu, S.~Xu, P.~Xu, T.~Xiao, F.~Xia, J.~Wu, P.~Wohlhart, S.~Welker, A.~Wahid \emph{et~al.}, ``Rt-2: Vision-language-action models transfer web knowledge to robotic control,'' in \emph{Conference on Robot Learning}.\hskip 1em plus 0.5em minus 0.4em\relax PMLR, 2023, pp. 2165--2183.

\bibitem{liu2024lost}
N.~F. Liu, K.~Lin, J.~Hewitt, A.~Paranjape, M.~Bevilacqua, F.~Petroni, and P.~Liang, ``Lost in the middle: How language models use long contexts,'' \emph{Transactions of the association for computational linguistics}, vol.~12, pp. 157--173, 2024.

\bibitem{wu2022ai}
T.~Wu, M.~Terry, and C.~J. Cai, ``Ai chains: Transparent and controllable human-ai interaction by chaining large language model prompts,'' in \emph{Proceedings of the 2022 CHI conference on human factors in computing systems}, 2022, pp. 1--22.

\bibitem{warren1982efficient}
D.~H. Warren and F.~C. Pereira, ``An efficient easily adaptable system for interpreting natural language queries,'' \emph{American journal of computational linguistics}, vol.~8, no. 3-4, pp. 110--122, 1982.

\bibitem{freitas2014comprehensible}
A.~A. Freitas, ``Comprehensible classification models: a position paper,'' \emph{ACM SIGKDD explorations newsletter}, vol.~15, no.~1, pp. 1--10, 2014.

\bibitem{wang2022self}
X.~Wang, J.~Wei, D.~Schuurmans, Q.~Le, E.~Chi, S.~Narang, A.~Chowdhery, and D.~Zhou, ``Self-consistency improves chain of thought reasoning in language models,'' \emph{arXiv preprint arXiv:2203.11171}, 2022.

\bibitem{weinberg2019selecting}
A.~I. Weinberg and M.~Last, ``Selecting a representative decision tree from an ensemble of decision-tree models for fast big data classification,'' \emph{Journal of Big Data}, vol.~6, no.~1, p.~23, 2019.

\bibitem{zheng2023judging}
L.~Zheng, W.-L. Chiang, Y.~Sheng, S.~Zhuang, Z.~Wu, Y.~Zhuang, Z.~Lin, Z.~Li, D.~Li, E.~Xing \emph{et~al.}, ``Judging llm-as-a-judge with mt-bench and chatbot arena,'' \emph{Advances in neural information processing systems}, vol.~36, pp. 46\,595--46\,623, 2023.

\bibitem{croxford2025current}
E.~Croxford, Y.~Gao, N.~Pellegrino, K.~Wong, G.~Wills, E.~First, F.~Liao, C.~Goswami, B.~Patterson, and M.~Afshar, ``Current and future state of evaluation of large language models for medical summarization tasks,'' \emph{Npj health systems}, vol.~2, no.~1, p.~6, 2025.

\bibitem{li2025preference}
D.~Li, R.~Sun, Y.~Huang, M.~Zhong, B.~Jiang, J.~Han, X.~Zhang, W.~Wang, and H.~Liu, ``Preference leakage: A contamination problem in llm-as-a-judge,'' \emph{arXiv preprint arXiv:2502.01534}, 2025.

\bibitem{breiman2017classification}
L.~Breiman, J.~Friedman, R.~A. Olshen, and C.~J. Stone, \emph{Classification and regression trees}.\hskip 1em plus 0.5em minus 0.4em\relax Chapman and Hall/CRC, 2017.

\bibitem{canal2017taxonomy}
G.~Canal, G.~Aleny{\`a}, and C.~Torras, ``A taxonomy of preferences for physically assistive robots,'' in \emph{2017 26th IEEE international symposium on Robot and human interactive communication (RO-MAN)}.\hskip 1em plus 0.5em minus 0.4em\relax IEEE, 2017, pp. 292--297.

\bibitem{akalin2023taxonomy}
N.~Akalin, A.~Kiselev, A.~Kristoffersson, and A.~Loutfi, ``A taxonomy of factors influencing perceived safety in human--robot interaction,'' \emph{International Journal of Social Robotics}, vol.~15, no.~12, pp. 1993--2004, 2023.

\bibitem{koay2014social}
K.~L. Koay, D.~S. Syrdal, M.~Ashgari-Oskoei, M.~L. Walters, and K.~Dautenhahn, ``Social roles and baseline proxemic preferences for a domestic service robot,'' \emph{International Journal of Social Robotics}, vol.~6, no.~4, pp. 469--488, 2014.

\bibitem{choi2009hand}
Y.~S. Choi, T.~Chen, A.~Jain, C.~Anderson, J.~D. Glass, and C.~C. Kemp, ``Hand it over or set it down: A user study of object delivery with an assistive mobile manipulator,'' in \emph{RO-MAN 2009-the 18th IEEE international symposium on robot and human interactive communication}.\hskip 1em plus 0.5em minus 0.4em\relax IEEE, 2009, pp. 736--743.

\bibitem{steinfeld2009oz}
A.~Steinfeld, O.~C. Jenkins, and B.~Scassellati, ``The oz of wizard: simulating the human for interaction research,'' in \emph{Proceedings of the 4th ACM/IEEE international conference on Human robot interaction}, 2009, pp. 101--108.

\bibitem{norman2014some}
D.~A. Norman, ``Some observations on mental models,'' in \emph{Mental models}.\hskip 1em plus 0.5em minus 0.4em\relax Psychology Press, 2014, pp. 7--14.

\bibitem{kidd1995functional}
D.~Kidd, G.~Stewart, J.~Baldry, J.~Johnson, D.~Rossiter, A.~Petruckevitch, and A.~Thompson, ``The functional independence measure: a comparative validity and reliability study,'' \emph{Disability and rehabilitation}, vol.~17, no.~1, pp. 10--14, 1995.

\bibitem{woods2006methodological}
S.~N. Woods, M.~L. Walters, K.~L. Koay, and K.~Dautenhahn, ``Methodological issues in hri: A comparison of live and video-based methods in robot to human approach direction trials,'' in \emph{ROMAN 2006-the 15th IEEE international symposium on robot and human interactive communication}.\hskip 1em plus 0.5em minus 0.4em\relax IEEE, 2006, pp. 51--58.

\bibitem{esterwood2025virtually}
C.~Esterwood, R.~H. Guan, X.~Ye, and L.~P. Robert, ``Virtually the same or realistically different?: A meta-analysis of real vs.‘not so real’robots,'' in \emph{2025 20th ACM/IEEE International Conference on Human-Robot Interaction (HRI)}.\hskip 1em plus 0.5em minus 0.4em\relax IEEE, 2025, pp. 559--568.

\bibitem{irfan2018social}
B.~Irfan, J.~Kennedy, S.~Lemaignan, F.~Papadopoulos, E.~Senft, and T.~Belpaeme, ``Social psychology and human-robot interaction: An uneasy marriage,'' in \emph{Companion of the 2018 ACM/IEEE international conference on human-robot interaction}, 2018, pp. 13--20.

\bibitem{hart1988development}
S.~G. Hart and L.~E. Staveland, ``Development of nasa-tlx (task load index): Results of empirical and theoretical research,'' in \emph{Advances in psychology}.\hskip 1em plus 0.5em minus 0.4em\relax Elsevier, 1988, vol.~52, pp. 139--183.

\bibitem{laban2022summac}
P.~Laban, T.~Schnabel, P.~N. Bennett, and M.~A. Hearst, ``Summac: Re-visiting nli-based models for inconsistency detection in summarization,'' \emph{Transactions of the Association for Computational Linguistics}, vol.~10, pp. 163--177, 2022.

\bibitem{van2024field}
T.~A. Van~Schaik and B.~Pugh, ``A field guide to automatic evaluation of llm-generated summaries,'' in \emph{Proceedings of the 47th International ACM SIGIR Conference on Research and Development in Information Retrieval}, 2024, pp. 2832--2836.

\end{thebibliography}

\end{document}